# Inorganic Catalyst Efficiency Prediction Based on EAPCR Model: A Deep Learning Solution for Multi-Source Heterogeneous Data

Zhangdi Liu[1,#], Ling An[1,#] , Mengke Song[1,#] , Zhuohang Yu[1] , Shan Wang[2,*] , Kezhen Qi[2,*] , Zhenyu Zhang[1], and Chichun Zhou[1,*]

[1]School of engineering, Dali university; Air-Space-Ground Integrated Intelligence and Big Data Application Engineering Research Center of Yunnan Provincial Department of Education, Yunnan, 671003, China

[2]College of Pharmacy, Dali University, Dali 671000, China

[#] These authors contribute equivalently

[*] Corresponding author: Chichun Zhou: zhouchichun@dali.edu.cn

Kezhen Qi: qkzh2003@aliyun.com

Shan Wang: shanwang1216@foxmail.com

Email of authors: Zhangdi Liu: liuzhangdi@stu.dali.edu.cn

Ling An: anling@stu.dali.edu.cn

Mengke Song: songmengke@stu.dali.edu.cn

Zhuohang Yu: yuzhuohang@stu.dali.edu.cn

Zhenyu Zhang: zhangzhenyu@dali.edu.cn

# Abstract

The design of inorganic catalysts and the prediction of their catalytic efficiency are fundamental challenges in chemistry and materials science. Traditional catalyst evaluation methods primarily rely on machine learning techniques; however, these methods often struggle to process multi-source heterogeneous data, limiting both predictive accuracy and generalization. To address these limitations, this study introduces the Embedding-Attention-Permutated CNN-Residual (EAPCR) deep learning model. EAPCR constructs a feature association matrix using embedding and attention mechanisms and enhances predictive performance through permutated CNN architectures and residual connections. This approach enables the model to accurately capture complex feature interactions across various catalytic conditions, leading to precise efficiency predictions. EAPCR serves as a powerful tool for computational researchers while also assisting domain experts in optimizing catalyst design, effectively bridging the gap between data-driven modeling and experimental applications. We evaluate EAPCR on datasets from $TiO_2$ photocatalysis, thermal catalysis, and electrocatalysis, demonstrating its superiority over traditional machine learning methods (e.g., linear regression, random forest) as well as conventional deep learning models (e.g., ANN, NNs). Across multiple evaluation metrics (MAE, MSE, $R^2$, and RMSE), EAPCR consistently outperforms existing approaches. These findings highlight the strong potential of EAPCR in inorganic catalytic efficiency prediction. As a versatile deep learning framework, EAPCR not only improves predictive accuracy but also establishes a solid foundation for future large-scale model development in inorganic catalysis.

# 1 Introduction

Inorganic catalysis is a crucial pillar of the modern chemical industry, with a rich history and ongoing advancements (Schmidt et al., 2004). With technological progress, the types, properties, and application domains of catalysts have continually expanded. Inorganic catalysts (ICs), as a vital branch of catalysis, have become indispensable components in various fields such as the chemical industry, energy development, and environmental protection due to their unique catalytic properties and broad applications (Büchel et al., 2008). ICs mainly consist of inorganic compounds, such as metals, metal oxides, metal salts, silicon-based materials, and nitrides. They lower the activation energy of chemical reactions, significantly accelerate reaction rates, and improve selectivity and efficiency, thus enabling efficient chemical production and conversion (Thomas et al., 2014). Recently, with a focus on sustainable development and green chemistry, research and applications of IC have increasingly prioritized environmental sustainability, efficiency, and renewability. ICs have widespread applications in the chemical industry, with key industrial processes such as ammonia synthesis, methanol synthesis, and petroleum hydrocracking (PHC) all relying on their catalytic actions (Bao et al., 2024). These processes not only lay the foundation for chemical production but also play a significant role in enhancing production efficiency, reducing energy consumption, and improving product quality. The advancement of ICs has not only led to innovations in traditional industries but also promoted the development and application of new chemical products. In the energy sector, ICs play a pivotal role. For example, precious metals such as platinum, iron, cobalt, and nickel, as well as their oxides, carbides, and borides, significantly enhance the efficiency of redox reactions, thus improving the overall performance and lifespan of fuel cells (Senthil et al., 2024). Additionally, ICs are widely used in the conversion and storage of renewable energy sources, such as solar energy and wind energy, providing strong support for the clean and efficient use of energy (Molaei et al., 2024). In environmental protection, ICs also play a crucial role. For example, using ICs in wastewater treatment and air purification can effectively remove harmful substances and reduce environmental pollution. Furthermore, ICs can catalyze the decomposition of toxic compounds, thereby reducing their impact on both the environment and human health (Bhagat et al., 2023). Research on ICs is not confined to chemistry alone but also, closely intersects with materials science, nanotechnology, and biological sciences (Somorjai et al., 2010). By exploring catalytic mechanisms, reaction pathways, and the relationship between structure and performance, these studies advance related fields and provide novel approaches for future research. In conclusion, ICs, as a key components of the modern chemical industry, play a pivotal roles in driving industrial progress, energy development, environmental protection, and scientific innovation.

As IC technology advances, new catalytic materials and catalysts continue to emerge, making the factors influencing catalytic efficiency increasingly complex. Traditionally, the prediction of catalytic efficiency has relied heavily on the accumulation of experimental data and statistical analysis (Hattori et al., 1995). However, these conventional approaches have significant limitations. First, acquiring experimental data requires substantial time and resource investment, and the inability to fully control experimental conditions often leads to data uncertainty and measurement errors. Second, statistical analysis methods typically provide only averaged or generalized predictions, failing to capture the variations in catalytic efficiency under different conditions. As a result, traditional prediction methods struggle to meet the demand for precise and reliable catalytic efficiency forecasting.

During the implementation of the Materials Genome Initiative (MGL), some researchers have begun exploring advanced algorithms such as machine learning to predict inorganic catalytic efficiency (Himanen et al., 2019). By analyzing large volumes of experimental data, these algorithms can automatically identify key factors influencing catalytic efficiency and develop predictive models. For example, Liu et al. (2022) employed traditional machine learning methods, such as Random Forest, to predict the degradation efficiency of photocatalytic titanium dioxide. Schossler et al. (2024) utilized Extreme Gradient Boosting (XGBoost) and CatBoost to estimate the degradation rate of titanium dioxide in photocatalysis. Similarly, Miyazato et al. (2020) applied linear regression (LR) and support vector machines (SVM) to predict the oxidative coupling of methane (OCM) reaction, while Chen et al. (2022) leveraged neural network (NN) models to predict adsorption energy in electrocatalysis. Although these approaches have improved the accuracy of catalytic efficiency predictions to some extent, they still face several challenges. Inorganic catalytic data often exhibit multi-source heterogeneity, encompassing variables such as temperature, pH, and dopants, with no explicit relationships among them. Traditional machine learning methods struggle to effectively capture these complex feature interactions, limiting their performance in catalytic efficiency prediction. More importantly, deep learning applications in this field remain relatively scarce. While deep learning has demonstrated remarkable success in other domains, its adoption in inorganic catalysis is still in its early stages due to the inherent complexity of catalytic data. Thus, a critical challenge remains: how to leverage deep learning effectively to uncover hidden relationships across multi-source data.

To address the aforementioned challenges, this study introduces a deep learning-based approach called EAPCR (Embedding-Attention-Permutated CNN-Residual), designed to efficiently integrate multi-source heterogeneous data and achieve accurate catalytic efficiency predictions. To validate its effectiveness, we conducted experiments on datasets from three catalytic domains: titanium dioxide photocatalysis, thermal catalysis, and electrocatalysis. The results demonstrate that EAPCR not only delivers precise efficiency predictions but also maintains stable performance across various catalytic environments, providing domain experts with a powerful tool for more accurate catalytic efficiency assessments. EAPCR leverages Embedding and Attention mechanisms to construct a feature association matrix, automatically capturing complex relationships within multi-source heterogeneous data—something traditional machine learning methods typically require manual feature engineering to achieve. By integrating Permutated CNN, EAPCR efficiently extracts critical features from the matrix, while the Residual Block enhances the model's stability and expressive power. This end-to-end feature learning approach minimizes the need for manual intervention, allowing method experts to focus on model optimization and innovation, ultimately improving model applicability. As a general deep learning framework, EAPCR bridges the gap between method experts and domain experts, offering robust technological support for inorganic catalytic efficiency prediction. Furthermore, it lays a solid foundation for the development of large-scale catalytic models in the future.

The main contributions of this paper are as follows:

- The EAPCR method introduced in this study integrates Embedding, Attention, Permutated CNN, and Residual modules to efficiently fuse multi-source heterogeneous data and extract complex feature relationships, providing a novel solution for predicting inorganic catalytic efficiency.
- EAPCR bridges the gap between methodology experts and domain specialists. By constructing an end-to-end automated learning framework, it can automatically learn complex relationships within the data while reducing human intervention. This enables methodology experts to focus more on model optimization while providing domain specialists with a more intuitive and efficient predictive tool.
- The advantages of deep learning in predicting inorganic catalysis efficiency are validated. Through experiments in photocatalysis, thermocatalysis, and electrocatalysis, the EAPCR method outperforms traditional statistical analysis and machine learning approaches in key performance metrics such as the Mean Absolute Error (MAE), Mean Squared Error (MSE), $R^2$, and Root Mean Squared Error (RMSE).

This provides strong support for the design and optimization of catalysts.

# 2 Materials and Methodology

## 2.1 Methodology: a review of EAPCR

The EAPCR was proposed in the previous work (Yu et al., 2024). Here we give a review of this method. In this model, we construct the correlation matrix by embedding (Mikolov et al., 2013) and bilinear attention (Kim et al., 2018). For datasets with sparse features, we first convert each feature into a categorical (string-based) feature. Categorical features, such as the type of catalyst, remain unchanged, whereas numerical features are discretized into categories, on the basis of specific thresholds. The choice of these thresholds is aimed at striking a balance between granularity—too fine a granularity may lead to sparse categories, whereas too coarse a granularity may reduce the distinction between features. Therefore, careful consideration is required when selecting thresholds to maximize the separability of features. This process generates an input matrix of shape [N,1], where each element is an integer index that maps the categorical value to the corresponding index through a dictionary.

The core of the matrix module involves constructing the feature matrix for the multisource heterogeneous data through matrix inner-product operations. Specifically, the vectors representing the sequence of factors affecting the degradation rate are first passed through the embedding layer. The primary goal of this step is to map each feature value into a continuous space, making it easier for further computations and modeling.

$$E=Embedding(X) \tag{1}$$

Here, X represents the embedded representation of the factor sequence.

Next, we define bilinear attention to compute the interactions between different features. By employing bilinear attention (Kim et.al, 2018), we can capture the complex relationship patterns among features, thus enhancing the model's performance. The corresponding formula is as follows:

$$A=EE^T \tag{2}$$

Here, the matrix A has a shape of [N,N], representing the constructed correlation matrix. Matrix A is crucial because each element within it signifies the relationship between two features. Any combination of elements can reveal latent feature interaction patterns. Consequently, matrix A encodes all possible relationships between the features, which is essential for further analysis.

**Design of the Permutation Matrix M:** To identify feature combinations with strong interactions from matrix A, we apply Convolutional Neural Networks (CNNs) to sample matrix A (Bronstein et al., 2017). CNNs are particularly effective at capturing patterns in local regions, making them well suited for detecting local feature interactions. By increasing the size of the convolutional kernels or adding more layers to the network, we can expand the receptive field of the CNN, thereby enabling it to sample different elements from matrix A more effectively. To further optimize feature extraction, we design a permutation matrix M that reorders the elements of matrix A, adjusting the relative positions of the features. This operation ensures that originally adjacent elements are no longer next to each other, whereas elements that were originally farther apart are brought closer together. The core idea behind this operation is to create an invertible permutation that reshuffles the order of N elements (1, 2, 3..., N).

The process works as follows: First, we arrange the N elements into an R×L matrix, where N=R×L and R and L are approximately equal. Next, we transpose this matrix and reshape it into an N×1 column vector. This new sequence represents the transformed positions of the original data.

Afterward, we create a zero matrix M of size N×N and populate it with 1 s at the positions indicated by the reshaped sequence, forming the permutation matrix. For example, suppose N=9. We arrange the numbers 1 through 9 into a 3×3 matrix as follows:[[1,2,3]， [4,5,6] and [7,8,9]]. We then transpose and reshape the matrix into a new sequence [1,4,7,2,5,8,3,6,9], which is used to fill a new matrix, resulting in a permutation matrix. By doing so, the distance between adjacent elements in the new sequence is at least 3, meaning that previously adjacent elements are no longer adjacent, whereas originally distant elements are now brought closer together. Finally, we apply the permutation matrix MMM to matrix A to obtain a new matrix P:

$$P \triangleq MAM^T \tag{3}$$

**Permutation-based CNN:** To enhance feature extraction, we introduce a permutation-based CNN architecture. The main idea behind this approach is to leverage the CNN’s ability to capture both local and nonlocal interactions between the original matrix A and the permutation matrix P. Here, we apply the CNN separately to both matrices—A and P—using a lightweight setup, such as 3×3 convolutional filters with 8 and 16 channels. This allows us to extract features from both matrices, resulting in two distinct feature representations. Once we have these two feature vectors, we concatenate them and pass them through a fully connected layer to generate the final prediction vector. This process enables the model to integrate the information from both matrices in a meaningful way.

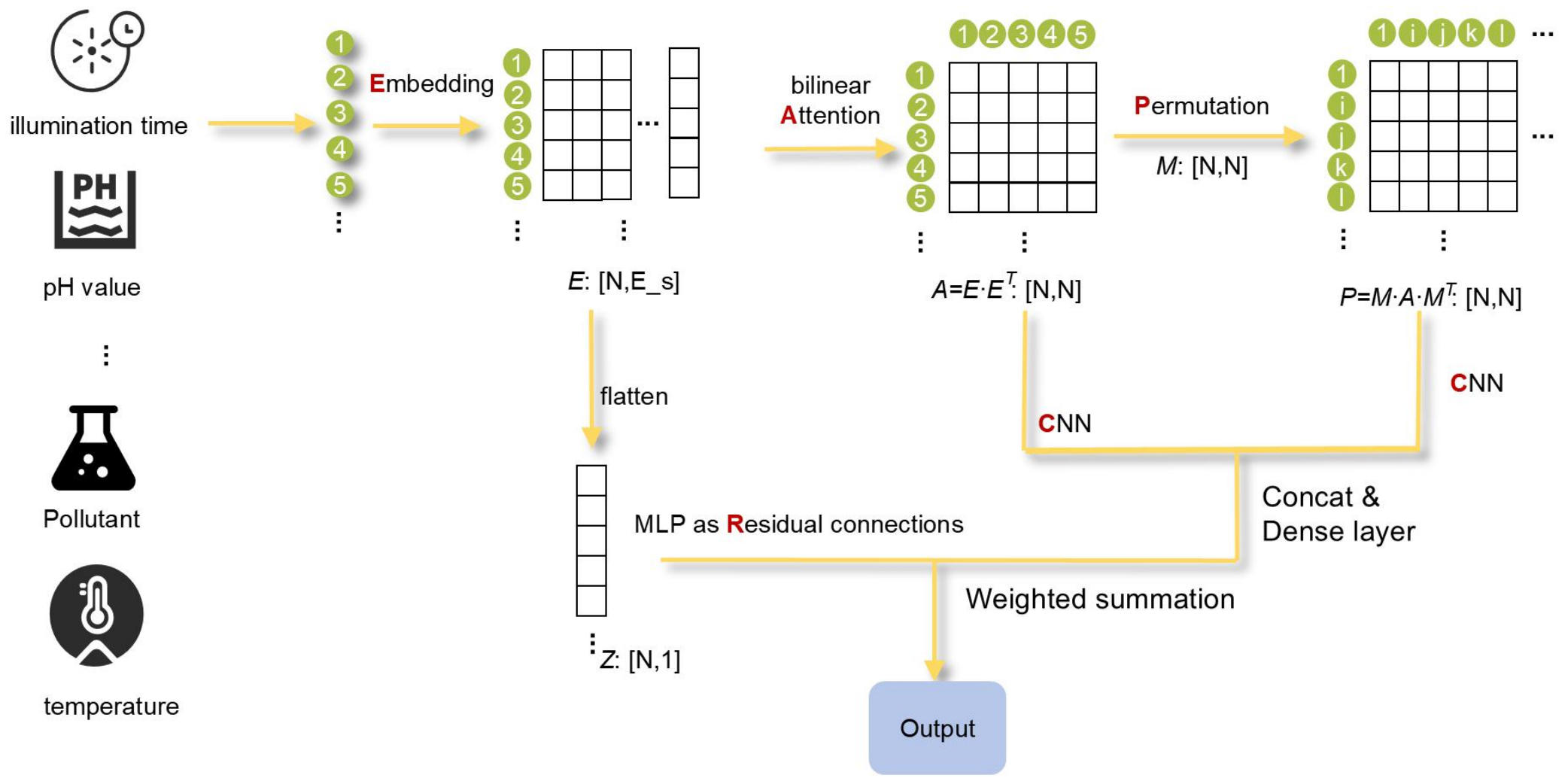

**Figure 1:** Overview of the EAPCR Model Architecture

In addition, we further improve the efficiency of the feature extraction process by introducing residual connections, specifically from modules such as the Multi-Layer Perceptron (MLP). To do this, we first flatten the output matrix E, converting it into a vector z, which is then processed by the MLP to produce the final prediction. These residual connections not only speed up the training process but also enhance the model's ability to learn the embedding vectors more effectively. This is all illustrated in Figure 1.

In this study, the prediction of IC efficiency is formulated as a regression problem. To assess the model's performance, we use the Mean Squared Error (MSE) as the loss function, The MSE metric helps to measure the model's accuracy by penalizing large errors, ensuring that the predicted values are as close as possible to the actual values. which is calculated as follows:

$$L_{MSE} = \frac{1}{N}\sum_{j=1}^{N}(\overline{y}_j - y_j)^2 \quad (4)$$

where $\overline{y_j}$ is the predicted value and where $y_j$ is the true value for the data point.

## 2.2 Database

To evaluate the effectiveness of the EAPCR method in predicting catalytic efficiency in the field of inorganic catalysis, we used publicly available datasets from three catalytic modes: photocatalysis, thermocatalysis, and electrocatalysis. These datasets are derived from published articles and can be downloaded from the referenced literature. An overview of the datasets is shown in Table 1.

**Table 1.** The number of datasets and the introduction of variables.

| | Number of instances | Variables | Target |
|---|---|---|---|
| Data1 | 760 | Dopant，Dopant/Ti mole ratio，Calcination temperature，Pollutant，Catalyst/Pollutant mass ratio，pH，Experimental temperature，Light wavelength，Illumination time， | Degradation rate |
| Data2 | 201 | OC，I mW/$cm^2$，W nm，D mg/$cm^2$，H %，T ℃，R L，Initial C ppmv | K, $min^{-1}/cm^2$ |
| Data3 | 375 | OC，I mW/$cm^2$，T ℃，D g/L，$C_0$ mg/L，pH | K, $min^{-1}$ |
| Data4 | 12708 | M1_atom_number，M2_atom_number，M3_atom_number，Support_ID，M1_mol%，M2_mol%，M3_mol%，Temp，Total_flow，CH4_flow，CT，CH4/O2 | CH4_conv |
| Data5 | 196 | Weight，CH4+O2，Total_flow，CH4O2，Temp | H2，CO，CO2，C2H6，C2H4 |
| Data6 | 2842 | M1，M2，M3，Support，Temperature，pinert，pch4，po2，CH4/O2 | C2y |
| Data7 | 280 | Site 1 Group，Site 1 Period，Site 1 EN，Site 1 Nied，...，Site 10 Group，Site 10 Period，Site 10 EN，Site 10 Nied | Labels |

**Dataset 1 (Liu et al., 2022):** This dataset contains experimental data for titanium dioxide ($TiO_2$) photocatalysis. Each data point includes 9 experimental variables: dopant type, dopant/Ti molar ratio, calcination temperature (℃), pollutant type, catalyst/pollutant mass ratio, pH, experimental temperature (℃), light wavelength (nm), and illumination time (min). The dopants consist of nonmetal elements (such as C, F, I, and N) as well as metal elements (such as Ag, Bi, Cd, and Ce). The pollutants include methyl blue, phenol, methyl orange, benzoic acid, and acid orange, among others. The dopant/Ti molar ratio ranges from 0 to 93:5, the calcination temperature varies from 400℃ to 900℃, and the catalyst/pollutant mass ratio ranges from 5:1 to 1000:1. The pH ranged from 2 to 13, while the experimental temperature, varied between 16℃ and 32℃. The light wavelength ranged from 254 nm to 600 nm, and the illumination time ranged from 5 to 480 minutes. The output variable is the degradation rate, which reflects the efficiency of pollutant removal during the catalytic process.

**Dataset 2 (Schossler et al., 2024):** This dataset is also used for titanium dioxide ($TiO_2$) photocatalysis and includes 8 experimental variables: type of organic pollutant (OC), ultraviolet light intensity (I, mW/$cm^2$), wavelength (W, nm), amount of $TiO_2$ (D,

mg/cm$^2$), humidity (H, %), experimental temperature (T, ℃), reactor volume (R, L), and initial pollutant concentration (InitialC, ppmv). The light intensity ranges from 0.36 to 75 mW/cm$^2$, the illumination wavelength spans from 253.7 to 370 nm, and the $TiO_2$ amount varies from 0.012 to 5.427 mg/cm$^2$. The humidity ranged from 0% to 1600%, while the experimental temperature varied between 22℃ and 350℃. The reactor volume ranged from 0.04 to 216 L, and the initial pollutant concentration ranged from 0.001 to 5944 ppmv. The response variable is the photodegradation rate (k, min$^{-1}$/cm$^2$), which reflects the efficiency of pollutant degradation under light exposure. For missing data, the K-Nearest neighbors method was used to estimate and fill in the missing values.

**Dataset 3 (Jiang et al., 2020):**This dataset, which is also used for titanium dioxide ($TiO_2$) photocatalysis, consists of 6 experimental variables: the type of organic pollutant (OC), ultraviolet light intensity (I, mW/cm$^2$), experimental temperature (T, ℃), $TiO_2$ dosage (D, gL$^{-1}$), initial concentration of water pollutants ($C_0$, mg/L), and the initial pH of the solution. The range of light intensity spans from 0.176 to 75 mW/cm$^2$, while the experimental temperature varies between 20℃ and 60℃. The amount of $TiO_2$ used in the experiments ranges from 0 to 7.5 gL$^{-1}$, and the concentration of water pollutants falls within the range of 0.13 to 342.47 mg/L. The pH value of the solution varied from 2 to 11. The output variable for this dataset is the photodegradation rate constant (k, min$^{-1}$), which indicates the rate at which pollutants are degraded under light exposure.

**Dataset 4 (Puliyanda et al., 2024):** This dataset focuses on the methane oxidative coupling (OCM) reaction and includes 12 experimental variables: the atomic numbers of three metal elements (M1_atom_number, M2_atom_number and M3_atom_number), the support material ID (Support_ID), the molar ratios of the metals (M1_mol%, M2_mol% and M3_mol%), the temperature (Temp), the volumetric flow rate (Total_flow), the methane flow rate ($CH_4$_flow), the reaction time (CT), and the methane-to-oxygen ($CH_4/O_2$) ratio. The output variable is the methane conversion rate, which measures the efficiency of methane conversion in the reaction.

**Dataset 5 (Miyazato et al., 2020):** This dataset is used for the methane oxidative coupling (OCM) reaction and includes five experimental variables: catalyst weight (mg), introduction amounts of methane and $O_2$ gases (conc%), total gas flow rate, the methane-to-$O_2$ ratio, and the temperature of the reaction tube (K). The response variable is the selectivity of five key products: $H_2$, $CO_2$, CO, $C_2H_6$, and $C_2H_4$, which indicates the selectivity of the products in the reaction.

**Dataset 6 (Nishimura et al., 2023):** This dataset is related to non-homogeneous

catalysis in the methane oxidative coupling (OCM) reaction and contains nine experimental variables: three metal elements (M1, M2, M3), catalyst support material (Support), temperature, volumetric flow rate (pinert), methane flow rate (pch4), oxygen flow rate (po2), and the methane-to-$O_2$ ratio ($CH_4/O_2$). The response variable is the C2 yield, which is a measure of the production of C2-series compounds during methane conversion.

**Dataset 7 (Chen et al., 2022):** This dataset focuses on predicting the adsorption energies of COOH*, CO, and CHO on high-entropy alloy (HEA) catalysts. It consists of 40 experimental variables, including different site positions (group, period, EN, etc.). The output variable is the adsorption energy, which quantifies the adsorption strength of various species on the catalyst surface.

# 3 Results

## 3.1 Main results

In this section, we present an analysis of the experimental results for Dataset 1. The training-to-test split of 7:3 was chosen to align with the dataset's source paper, ensuring the comparability and reliability of the results. The EAPCR model demonstrated impressive performance on this dataset, as shown in Table 2. Specifically, the model achieved an MAE of 0.128, an MSE of 0.041, an RMSE of 0.203, and an $R^2$ score of 0.937, significantly outperforming traditional methods such as linear regression, random forests, and XGBoost.

**Table 2:** Data 1 photocatalytic MSE, MAE, RMSE and $R^2$ indicators

| Model | MAE | MSE | RMSE | $R^2$ |
|---|---|---|---|---|
| Linear Regression (Liu et al., 2022) | 0.513 ± 0.104 | 0.601 ± 0.237 | 0.762 ± 0.401 | 0.048 ± 0.014 |
| RF (Liu et al., 2022) | 0.235 ± 0.062 | 0.180 ± 0.126 | 0.417 ± 0.148 | 0.805 ± 0.035 |
| XGBoost (Liu et al., 2022) | 0.145 ± 0.103 | 0.086 ± 0.034 | 0.293 ± 0.136 | 0.884 ± 0.024 |
| LightGBM (Liu et al., 2022) | / | / | / | 0.928 |
| **EAPCR** | **0.128 ± 0.003** | **0.041 ± 0.001** | **0.203 ± 0.004** | **0.937 ± 0.003** |

The $R^2$ score highlights the model's ability to capture the underlying data patterns effectively, with a strong correlation between the predicted and actual values. In contrast, traditional methods display considerable variability in the MAE and $R^2$ metrics, with linear regression exhibiting notable errors and relatively poor predictive performance. While random forests and XGBoost performed better than linear regression in some cases did, they still fell short of the accuracy and stability of the EAPCR model. This clearly demonstrates that the EAPCR model provides superior predictive performance and robustness for photocatalysis tasks. Figure 2 further illustrates the average prediction errors for the EAPCR model on each dataset. Each subplot displays the relationship between the predicted values (y-axis) and the true values (x-axis). The dashed line in the figures represents the ideal prediction line where y equals x, highlighting the accuracy of the model’s predictions. The scatter plot illustrating the model's performance on this dataset is shown in Figure 2-a.

**Table 3:** MSE, MAE, RMSE and $R^2$ indicators of photocatalysis in data 2.

| Model | MAE | MSE | RMSE | $R^2$ |
|---|---|---|---|---|
| ANN with BO (Schossler et al., 2024) | / | 0.559 | / | 0.438 |
| Catboost + Adaboost (Schossler et al., 2024) | / | 0.064 | / | 0.922 |
| GBM with BO (Schossler et al., 2024) | / | 0.117 | / | 0.882 |
| XGB with HYPEROPT (Schossler et al., 2024) | / | 0.073 | / | 0.927 |
| **EAPCR** | **0.131 ± 0.004** | **0.054 ± 0.001** | **0.233 ± 0.003** | **0.940 ± 0.002** |

In this section, we analyze the experimental results from Dataset 2, which was split into a 7:3 training-to-testing ratio to maintain consistency with the dataset in the original study. The performance of the EAPCR model on this dataset is equally impressive, as shown in Table 3. Specifically, the model achieved an MAE of 0.131, MSE of 0.054, RMSE of 0.233, and an $R^2$ score of 0.940, further demonstrating its superior performance in photocatalysis tasks. A high $R^2$ score indicates that the EAPCR model effectively captures the distribution patterns in the data, maintaining a strong correlation with the true values.

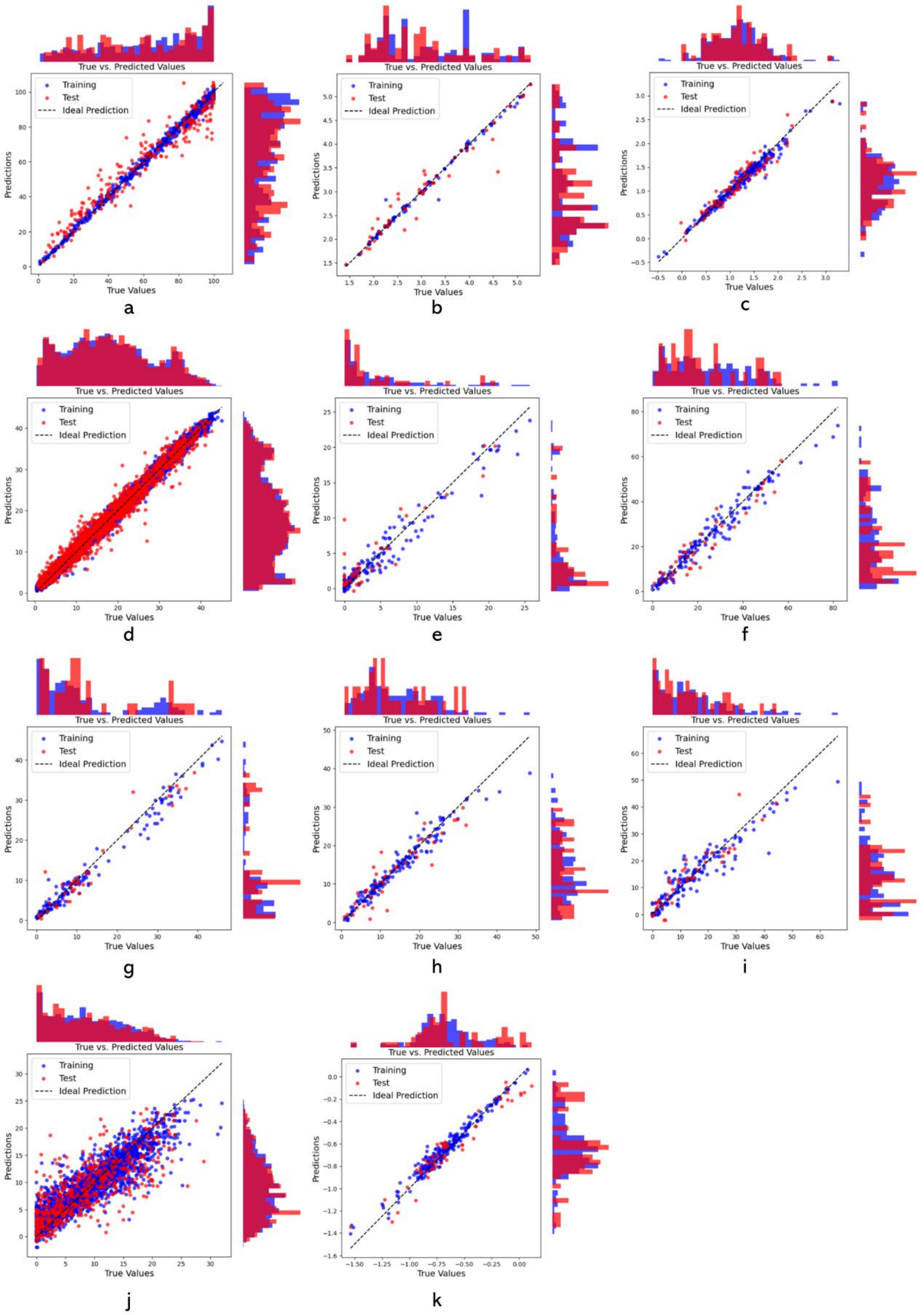

**Figure 2:** Scatter plots of model performance across different datasets. (a) Dataset 1, (b) Dataset 2, (c) Dataset 3, (d) Dataset 4, (e) Dataset 5: $H_2$ electivity, (f) Dataset 5: CO selectivity, (g) Dataset 5: $C_2H_6$ selectivity, (h) Dataset 5: $C_2H_4$ selectivity, (i) Dataset 5: $C_2H_4$ selectivity (repeated for clarity), (j) Dataset 6, (k) Dataset 7.

In contrast, the traditional methods exhibited similar performance trends as those observed in Dataset 1, with substantial fluctuations in prediction errors. For example, linear regression showed significantly higher error values (such as MAEs and MSEs) than did EAPCR, suggesting its limited ability to model photocatalysis data. While random forest and XGBoost performed somewhat better, they still fell short of EAPCR in terms of overall accuracy and generalizability capability. An analysis of scatter plot (Figure 2-b), clearly reveals that the EAPCR model not only has strong predictive power but also excels in capturing the complex characteristics of photocatalysis.

In the following section, we analyze the experimental results for Datasets 3 and 4. For Dataset 3, we used five-fold cross-validation for both training and testing. The EAPCR model demonstrated strong performance on this dataset, achieving an MAE of 0.121, an MSE of 0.029, an RMSE of 0.169, and an $R^2$ score of 0.905 (as shown in Table 4). These results indicate that the model offers high predictive accuracy, especially with its low error values, confirming its effectiveness for photocatalysis tasks. Figure 2-c shows the performance scatter plot for Dataset 3, where the predicted values closely follow the ideal line y = x, suggesting that the model fits the data features well.

**Table 4:** MSE, MAE, RMSE and $R^2$ values of data from 3 photocatalysis versus dataset 4 thermal catalysts.

| Data | Model | MAE | MSE | RMSE | $R^2$ |
|---|---|---|---|---|---|
| Data3 | Bayesian optimization of ANN topology (Jiang et al., 2020) | **0.108** | / | 0.173 | 0.873 |
| | **EAPCR** | 0.121 ± 0.002 | **0.029 ± 0.000** | **0.169 ± 0.002** | **0.905 ± 0.002** |
| Data4 | RF (Puliyanda et al., 2024) | / | / | 3.40 | 0.89 |
| | **EAPCR** | **1.275 ± 0.024** | **2.886 ± 0.096** | **1.698 ± 0.027** | **0.973 ± 0.000** |

On the other hand, for Dataset 4, we adopted a training-to-testing ratio of 8.5:1.5. While the error values were somewhat higher than those for Dataset 3, the EAPCR model still performed impressively, with an MAE of 1.275, an MSE of 2.886, an RMSE of 1.698, and an $R^2$ score of 0.973 (as presented in Table 4). Despite the increased error values, the model's $R^2$ score remains close to 1, indicating excellent fit and accuracy. The scatter plot of model performance for Dataset 4, shown in Figure 2-d, highlights the strong correlation between the predicted and actual values, further confirming the robustness and stability of the EAPCR model across different datasets.

For Dataset 5, the experiment was conducted with an 8:2 training-to-testing ratio, and a comprehensive evaluation was carried out across five selectivity labels. The detailed results are presented in Table 5 and Figures 2-e to 2-i. The EAPCR model showed excellent predictive performance across all the targets, particularly excelling in the selectivity predictions for $H_2$ and CO, with $R^2$ scores of 0.839 and 0.909, respectively. The scatter plots indicate that most data points are closely aligned with the ideal prediction line, demonstrating the model's high accuracy and reliability across different selectivity ranges.

**Table 5:** MSE, MAE, RMSE and $R^2$ metrics for thermal catalysis for dataset 5.

| Model | Target | MAE | MSE | RMSE | $R^2$ |
|---|---|---|---|---|---|
| RFR (Miyazato et al., 2020) | $H_2$ selectivity | / | / | / | 0.71 |
| **EAPCR** | | **1.472 ± 0.001** | **4.865 ± 0.009** | **2.205 ± 0.002** | **0.839 ± 0.000** |
| RFR (Miyazato et al., 2020) | CO selectivity | / | / | / | 0.80 |
| **EAPCR** | | **3.570 ± 0.020** | **22.048 ± 0.147** | **4.695 ± 0.015** | **0.909 ± 0.000** |
| RFR (Miyazato et al., 2020) | $C_2H_6$ selectivity | / | / | / | 0.83 |
| **EAPCR** | | **1.600 ± 0.049** | **6.005 ± 0.479** | **2.448 ± 0.099** | **0.954 ± 0.003** |
| RFR (Miyazato et al., 2020) | $CO_2$ selectivity | / | / | / | 0.77 |
| **EAPCR** | | **2.416 ± 0.005** | **11.595 ± 0.080** | **3.405 ± 0.011** | **0.843 ± 0.001** |
| RFR (Miyazato et | $C_2H_4$ selectivity | / | / | / | 0.77 |
| al., 2020) | | | | | |
| **EAPCR** | | **3.099 ± 0.007** | **16.273 ± 0.019** | **4.034 ± 0.002** | **0.868 ± 0.000** |

Moreover, the model performed exceptionally well in predicting $C_2H_6$ selectivity, achieving an $R^2$ score of 0.954, highlighting its impressive fit and stability. For $CO_2$ and $C_2H_4$ selectivity, the model achieved $R^2$ scores of 0.843 and 0.868, respectively, further confirming its broad applicability and outstanding performance in multi-target prediction. These results underscore the strong potential and practical value of the EAPCR model in complex catalytic systems.

For Dataset 6, the training-to-testing ratio was set at 8:2. The performance metrics of the EAPCR model were as follows: an MAE of 2.602, an MSE of 12.916, an RMSE of 3.594, and an $R^2$ score of 0.692. In comparison, the model showed significantly improved performance on Dataset 7, with an MAE of 0.076, an MSE of 0.009, an RMSE of 0.095, and an $R^2$ score of 0.924 (as shown in Table 6). The model performance scatter plots for Datasets 6 and 7 are presented in Figures 2-j and 2-k, respectively. These plots clearly visualize the accuracy and stability of the EAPCR model across both datasets.

**Table 6:** MSE, MAE, RMSE and $R^2$ indicators for thermal catalysis in Data 6 versus electric catalysis in Data 7.

| Data | Model | MAE | MSE | RMSE | $R^2$ |
|---|---|---|---|---|---|
| Data6 | SVR (Nishimura et al., 2023) | / | / | / | 0.54 |
| | **EAPCR** | **2.602 ± 0.007** | **12.916 ± 0.043** | **3.594 ± 0.006** | **0.692 ± 0.001** |
| Data7 | NN models (Chen et al., 2022) | 0.121 | 0.021 | / | / |
| | **EAPCR** | **0.076 ± 0.001** | **0.009 ± 0.001** | **0.095 ± 0.001** | **0.924 ± 0.002** |

## 3.2 Ablation study results

We analyzed the impact of different embedding sizes on the performance of the EAPCR model, specifically focusing on the $R^2$ score, with the detailed results provided in Table 7. The dataset split for training and testing remained consistent with that used in previous experiments. For Dataset 5, which contains multiple target variables, selectivity was chosen as the label for this analysis. A comparison of the results across different embedding sizes clearly demonstrates the significant effect of this hyperparameter on model performance. With smaller embedding sizes (such as 8 and 16), the model's performance was relatively lower for some datasets. However, as the embedding size increased, the model's performance improved steadily. For example, in Dataset 6, the highest $R^2$ score of 0.6925 was achieved with an embedding size of 64, whereas in Dataset 5, the best performance of 0.9544 was obtained with an embedding size of 128. Overall, appropriately adjusting the embedding size is crucial for enhancing the model's performance.

As shown in the results of Table 7, the EAPCR model's performance generally improves with an increasing embedding size. However, the benefits of enlarging the embedding size are not unlimited. While a larger embedding size can capture more of the original features and carry additional useful information, it may also introduce redundancy, increase computational costs, and increase the model's complexity. This overcomplication can negatively impact the model's training efficiency and reduce its generalization ability. Therefore, when selecting the embedding size in practical applications, it is essential to strike a balance between the amount of information and computational complexity, ensuring that performance and efficiency are optimized for the best model outcomes.

**Table 7**. shows the $R^2$ scores of the EAPCR model under different embedding_sizes.

| Embed_size | 8 | 16 | 32 | 64 | 128 | 256 |
|---|---|---|---|---|---|---|
| Data1 | 0.8108 | 0.8748 | 0.8909 | 0.9008 | 0.9318 | **0.9367** |
| Data2 | 0.8567 | 0.9387 | 0.9117 | 0.9137 | **0.9403** | 0.9146 |
| Data3 | 0.9103 | 0.9062 | 0.9315 | 0.9232 | 0.9328 | **0.9352** |
| Data4 | 0.9167 | 0.9249 | 0.9283 | 0.9214 | **0.9713** | 0.9691 |
| Data5 | 0.8608 | 0.9326 | 0.9333 | 0.9482 | **0.9544** | 0.9450 |
| Data6 | 0.6148 | 0.6112 | 0.6865 | **0.6925** | 0.6810 | 0.6642 |
| Data7 | 0.9095 | **0.9267** | 0.9223 | 0.9172 | 0.9129 | 0.9016 |

For datasets with sparse features, we first transform the features into categorical variables. Categorical features, such as catalyst type, remain unchanged, whereas numerical features are discretized into categories on the basis of specific thresholds (e.g., “high,” “medium,” “low”). The choice of thresholds must strike a balance between too-fine-grained categories (leading to sparsity) and too coarse-grained categories (which can reduce the discriminative power of the features). For example, temperature can be divided into five categories: “very high,” “high,” “medium,” “low,” and “very low.” To accomplish this, we used the Sklearn KBinsDiscretizer method, applying equal frequency binning to categorize the features on the basis of their distribution, ensuring that each category contains an equal number of samples. For example, the temperature feature can be divided into three categories: “low,” “medium,” and “high.” Table 8 illustrates the impact of different binning thresholds on model performance via equal frequency binning, with data from Dataset 6.

**Table 8.** Performance of models with different thresholds for Data6 equal frequency binning.

| | n_bins=8 | n_bins=14 | n_bins=20 | n_bins=28 | n_bins=30 | n_bins=36 |
|---|---|---|---|---|---|---|
| $R^2$ | 0.6403 | 0.6671 | 0.6553 | 0.6691 | **0.6925** | 0.6717 |
| MAE | 2.8622 | 2.6920 | 2.7488 | 2.6650 | **2.6026** | 2.6537 |
| MSE | 15.1118 | 13.9847 | 14.4824 | 13.9023 | **12.9168** | 13.7907 |
| RMSE | 3.8874 | 3.7396 | 3.8056 | 3.7286 | **3.5940** | 3.7136 |

As shown in the data in Table 8, the impact of different binning thresholds on model performance is significant. The general trend is that as the number of bins increases, the model's performance improves. However, when the number of bins becomes too large, the performance begins to decline. This is because discretizing numerical features into categories results in a substantial change in how the features are represented. Coarse binning (e.g., n_bins = 8) may fail to capture sufficient feature information, limiting the model's ability to distinguish between different patterns. On the other hand, excessively fine binning (e.g., n_bins = 36) can lead to sparse features, making it more difficult for the model to learn effectively. Therefore, when discretizing features, it is crucial to find a balance in the granularity, ensuring that the feature information is adequately expressed without introducing sparsity, which could hinder the model's learning process. For example, in the table, when n_bins = 30, the model achieves the highest $R^2$ of 0.6925, with MAE, MSE, and RMSE values of 2.602, 12.916, and 3.594, respectively, demonstrating optimal model performance.

# 4 Conclusion and discussion

In this study, we introduce the deep learning-based EAPCR method for predicting IC efficiency, particularly in practical applications such as photocatalysis, thermocatalysis, and electrocatalysis. While traditional machine learning methods (such as linear regression and random forests) have been widely applied in catalyst efficiency prediction, they often fail to effectively capture the deep relationships between different reaction conditions because of the complex and multi-source heterogeneous nature of inorganic catalytic data. Moreover, the application of deep learning models in this field remains relatively rare, highlighting the need for more advanced methods to bridge this gap.

The EAPCR method leverages deep learning algorithms to efficiently capture the complex feature correlations between catalytic reaction conditions, thereby improving prediction accuracy. Compared with classical neural networks (such as ANNs and NNs) and traditional machine learning methods, EAPCR not only significantly enhances prediction accuracy but also has stronger generalizability and adaptability. Specifically, we validated the performance of the EAPCR method on representative datasets from photocatalysis, thermocatalysis, and electrocatalysis. The results show that EAPCR outperforms traditional prediction methods, highlighting its clear advantages as a reliable predictive tool. More importantly, as a general deep learning model, EAPCR not only optimizes the existing prediction framework but also provides technical support for the future development of large-scale catalytic models, driving technological progress in this field.

In summary, the EAPCR method has demonstrated significant potential in predicting catalyst efficiency and offers an innovative technical pathway for IC research. With further optimization and application of the EAPCR method, we anticipate that it will advance IC research to new heights, facilitating more groundbreaking discoveries and innovations.

# Data and Code Availability

The public datasets can be found in the corresponding references. The source code and private dataset can be made available upon reasonable request to the corresponding author.

# Acknowledgments

This work was supported by the National Natural Science Foundation of China (62106033, 42367066), Yunnan Fundamental Research Projects (202401AT070016, 202301BA070001-037), Yunnan Province Dali Prefecture Science and Technology Bureau Social Development Field Project (20232904E030002).

# Author contribution statement

Conceptualization, Chichun Zhou, Shan Wang, Kezhen Qi, and Zhenyu Zhang; Data curation, Zhangdi Liu; Formal analysis, Chichun Zhou, Shan Wang, Kezhen Qi, and Zhenyu Zhang; Funding acquisition, Chichun Zhou and Zhenyu Zhang; Investigation,

Zhangdi Liu, Shan Wang, and Ling An; Methodology, Zhuohang Yu, Ling An, and Chichun Zhou; Validation, Zhangdi Liu, Mengke Song, and Ling An ;Writing-original draft, Zhangdi Liu, Ling An, Mengke Song, Shan Wang, and Chichun Zhou; Writing-review & editing, Chichun Zhou, Kezhen Qi, and Zhenyu Zhang. Supervision: Chichun Zhou, Kezhen Qi, and Zhenyu Zhang.

# Additional information

Conflicts of Interest: The authors declare that they have no known competing financial interests or personal relationships that could have appeared to influence the work reported in this paper.

# References

Bao, W., Zhang, C., Yan, D., Sun, F., Yue, C., Wang, C., & Lu, Y. (2024). The organic–inorganic dual anchoring strategy based on organic chelators and lacunary polyoxometalates for the preparation of efficient CoWS hydrodesulfurization catalysts. *Fuel*, 360, 130599.

Bhagat, S. K., Pilario, K. E., Babalola, O. E., Tiyasha, T., Yaqub, M., Onu, C. E., ... & Yaseen, Z. M. (2023). Comprehensive review on machine learning methodologies for modeling dye removal processes in wastewater. *Journal of Cleaner Production*, 385, 135522.

Büchel, K. H., Moretto, H. H., & Werner, D. (2008). *Industrial inorganic chemistry*. John Wiley & Sons.

Bronstein, M. M., Bruna, J., LeCun, Y., Szlam, A., & Vandergheynst, P. (2017). Geometric deep learning: going beyond Euclidean data. *IEEE Signal Processing Magazine*, 34(4), 18-42.

Chen, Z. W., Gariepy, Z., Chen, L., Yao, X., Anand, A., Liu, S. J., ... & Singh, C. V. (2022). Machine-learning-driven high-entropy alloy catalyst discovery to circumvent the scaling relation for CO2 reduction reaction. *ACS Catalysis*, 12(24), 14864-14871.

Hattori, H. (1995). Heterogeneous basic catalysis. *Chemical Reviews*, 95(3), 537-558.

Himanen, L., Geurts, A., Foster, A. S., & Rinke, P. (2019). Data-driven materials science: status, challenges, and perspectives. *Advanced Science*, 6(21), 1900808.

Jiang, Z., Hu, J., Zhang, X., Zhao, Y., Fan, X., Zhong, S., ... & Yu, X. (2020). A generalized predictive model for TiO2–Catalyzed photo-degradation rate constants of water contaminants through artificial neural network. *Environmental Research*, 187, 109697.

Kim, J. H., Jun, J., & Zhang, B. T. (2018). Bilinear attention networks. *Advances in*

*Neural Information Processing Systems*, 31.

Liu, Q., Pan, K., Lu, Y., Wei, W., Wang, S., Du, W., ... & Zhou, Y. (2022). Data-driven for accelerated design strategy of photocatalytic degradation activity prediction of doped TiO2 photocatalyst. *Journal of Water Process Engineering*, 49, 103126.

Mikolov, T. (2013). Efficient estimation of word representations in vector space. *arXiv preprint arXiv:1301.3781*.

Miyazato, I., Nishimura, S., Takahashi, L., Ohyama, J., & Takahashi, K. (2020). Data-driven identification of the reaction network in oxidative coupling of the methane reaction via experimental data. *The Journal of Physical Chemistry Letters*, 11(3), 787-795.

Molaei, M. J. (2024). Recent advances in hydrogen production through photocatalytic water splitting: A review. *Fuel*, 365, 131159.

Nishimura, S., Li, X., Ohyama, J., & Takahashi, K. (2023). Leveraging machine learning engineering to uncover insights into heterogeneous catalyst design for oxidative coupling of methane. *Catalysis Science & Technology*, 13(16), 4646-4655.

Puliyanda, A. (2024). Model-based catalyst screening and optimal experimental design for the oxidative coupling of methane. *Digital Chemical Engineering*, 100160.

Schmidt, F. (2004). The importance of catalysis in the chemical and non-chemical industries. *In Basic principles in applied catalysis* (pp. 3-16). Berlin, Heidelberg: Springer Berlin Heidelberg.

Schossler, R. T., Ojo, S., Jiang, Z., Hu, J., & Yu, X. (2024). A novel interpretable machine learning model approach for the prediction of TiO2 photocatalytic degradation of air contaminants. *Scientific Reports*, 14(1), 13070.

Somorjai, G. A., & Li, Y. (2010). *Introduction to surface chemistry and catalysis*. John Wiley & Sons.

Thomas, J. M., & Thomas, W. J. (2014). *Principles and practice of heterogeneous catalysis*. John Wiley & Sons.

Yu, Z., An, L., Li, Y., Wu, Y., Dong, Z., Liu, Z., ... & Zhou, C. (2024). EAPCR: A Universal Feature Extractor for Scientific Data without Explicit Feature Relation Patterns. *arxiv preprint arxiv*:2411.08164.

# Appendix

## Evaluation Metrics

In the performance evaluation phase, we comprehensively analyzed the predictive performance of the EAPCR method via multiple metrics. Specifically, we employed the Mean Absolute Error (MAE), Mean Squared Error (MSE), Root Mean Squared Error (RMSE), and the Coefficient of Determination ($R^2$) to assess the prediction errors and the model fit.

First, the Mean Absolute Error (MAE) represents the average absolute difference between the predicted and actual values, and is calculated as:

$$MAE = \frac{1}{n}\sum_{i=1}^{n}\left|y_i - \hat{y}_i\right| \tag{1}$$

where $y_i$ is the actual value, $\hat{y}_i$ is the predicted value, and n is the sample size. A lower MAE indicates better model performance with smaller prediction errors.

Second, the Mean Squared Error (MSE) measures the average squared difference between the predicted and actual values, and is calculated as:

$$MSE = \frac{1}{n}\sum_{i=1}^{n}\left(y_i - \hat{y}_i\right)^2 \tag{2}$$

Similar to the MAE, a smaller MSE suggests better model fit, as it reflects reduced error in the model's predictions.

The Root Mean Squared Error (RMSE) is the square root of the MSE and indicates the standard deviation of the prediction errors. It is calculated as:

$$RMSE = \sqrt{MSE} \tag{3}$$

A smaller RMSE indicates better alignment between the predicted and actual values, and thus a better model fit.

Finally, the Coefficient of Determination ($R^2$) is used to evaluate how well the model fits the data, and is calculated as follows:

$$R^2 = 1 - \frac{SSR}{SST} \tag{4}$$

where SSR (Sum of Squared Residuals) represents the residual sum of squares, and SST (Total Sum of Squares) is the total sum of squares. $R^2$ ranges from 0 to 1, with values closer to 1 indicating a better model fit, and values closer to 0 suggesting poorer performance.